\documentclass[letterpaper]{article} 
\usepackage{aaai2027}  
\usepackage[hyphens]{url}  
\usepackage{graphicx} 
\usepackage{natbib}  
\usepackage{caption} 
\usepackage{booktabs}
\usepackage{amsmath}
\usepackage{amssymb}
\usepackage{multirow}

\newcommand{\best}[1]{\textbf{#1}}

\providecommand{\up}{\ensuremath{\uparrow}}
\providecommand{\down}{\ensuremath{\downarrow}}

\usepackage{xcolor}     
\usepackage{colortbl}   

\usepackage{seqsplit}

\title{Skip the Talk, Re-Focus on Vision: Latent Reasoning for Reasoning Segmentation in Multimodal Large Language Models}

\author{
  Tianhang Guo$^{1,*}$ \quad
  Yulin He$^{1,*}$ \quad
  Wei Chen$^{1,\dagger}$ \\[0.3em]
  Wenjuan Zhou$^{1}$ \quad
  Yuhang Li$^{1}$ \quad
  Xinbiao Gan$^{1}$
}
\affiliations{
  $^{1}$National University of Defense Technology, Changsha, China\\
  $^{*}$Equal contribution.\quad
  $^{\dagger}$Corresponding author.\\
  Email: \texttt{guotianhang.\_@nudt.edu.cn}
}

\begin{document}
\maketitle
\begin{abstract}
Reasoning segmentation aims to interpret implicit textual queries and enable fine-grained visual perception, which is critical for applications such as human–computer interaction and embodied agents. Existing methods typically generate explicit Chain-of-Thought (CoT) by multimodal large language models (MLLMs) before localizing the target. Although intuitive, such explicit verbal reasoning introduces substantial attention interference: redundant textual tokens disrupt attention during perception-token generation and also increase the effective distance between visual tokens.
To address this issue, we propose LIRSeg, which fully replaces explicit CoT with a compact set of learnable latent tokens for reasoning segmentation. 
LIRSeg is trained in two stages: spatial alignment grounds the latent tokens in object-relevant visual evidence, and GRPO further optimizes them with segmentation rewards.
To make these compact latent tokens more informative, we introduce three complementary mechanisms from an information perspective: extreme-advantage sampling for selecting informative training signals, decoupled exploration–stability updates for learning complementary representations, and latent diversity amplification for preventing representational collapse.
Extensive experiments on benchmarks demonstrate that LIRSeg consistently improves both segmentation accuracy and reasoning efficiency. Compared with the VisionReasoner baseline, LIRSeg achieves absolute gIoU improvements of 4.9\% on ReasonSeg, 7.1\% on MUSE, and 4.7\% on MMR, while achieving a $\sim\!16\times$ reduction in reasoning tokens. Code is available in supplementary materials.
\end{abstract}
\section{Introduction}
Multimodal large language models (MLLMs)~\citep{llava,qwen25vl} unify language
generation with visual perception, achieving performance on open-world image--text understanding.
Beyond recognizing visual content, reasoning segmentation~\citep{lisa} demands a perception capability from MLLMs: given an image and a free-form query, models need to infer the intent and localize the target at the pixel level.
Unlike referring segmentation, where the target is explicitly named, queries in reasoning segmentation are implicit and require the model to integrate functional, contextual, and intentional cues to identify objects.

\begin{figure}[t]
\centering
\includegraphics[width=0.99\columnwidth]{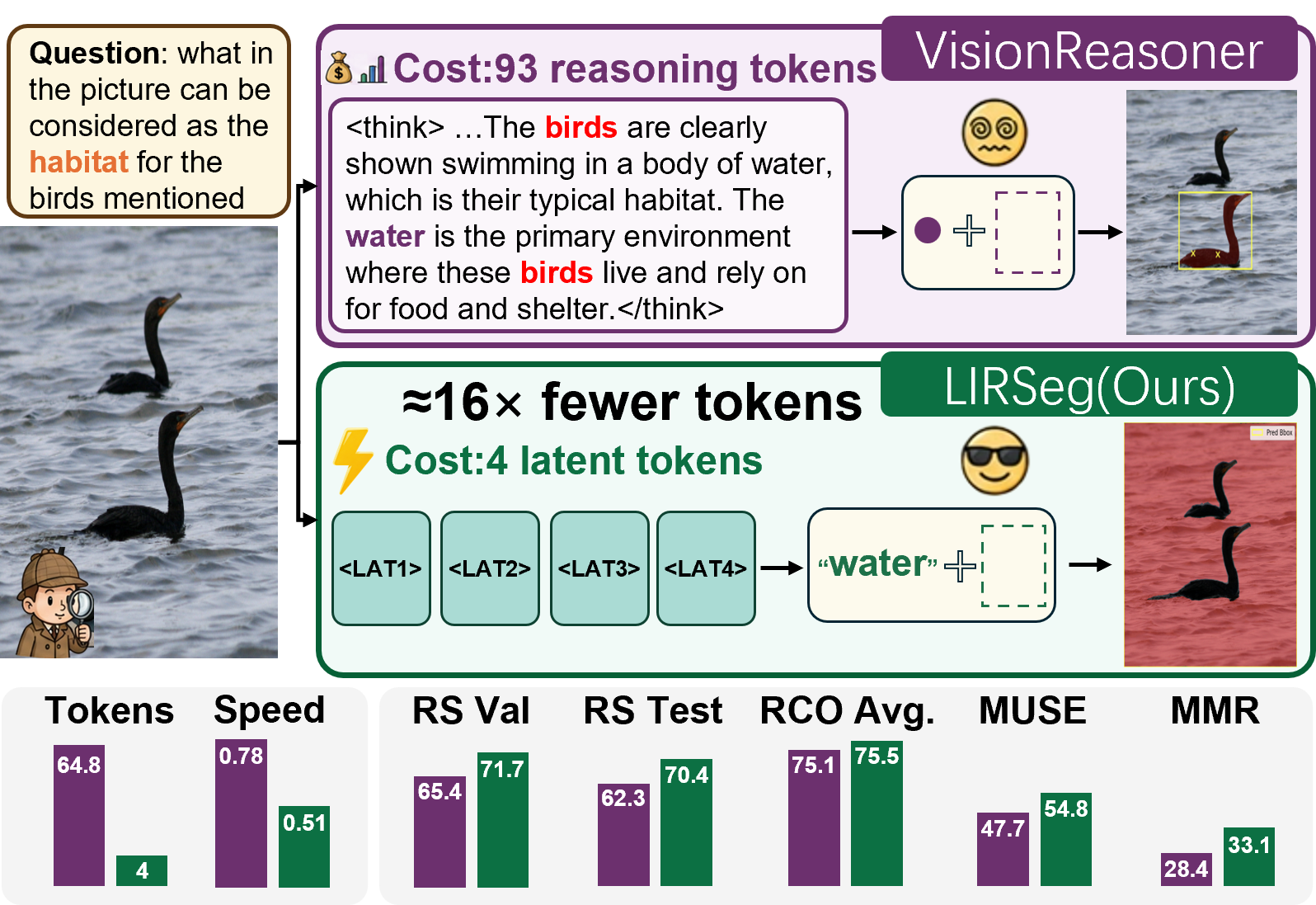}
\caption{\textbf{Motivation.} LIRSeg implicitly reasons with few latent tokens, cutting reasoning tokens by $\sim\!16\times$ while improving accuracy on ReasonSeg, RefCOCO, MUSE, and MMR, unlike explicit-CoT methods that distract from targets.}
\label{fig:teaser}
\end{figure}

Existing methods broadly fall into two paradigms. Supervised fine-tuning (SFT) methods, represented by LISA~\citep{lisa}, combine an MLLM with a segmentation model and train on annotated data. While effective in-distribution, they often generalize poorly to out-of-distribution (OOD) due to limited reasoning traces in annotated data. Reinforcement learning (RL) methods, such as Seg-Zero~\citep{segzero}, instead optimize the MLLM with Group Relative Policy Optimization (GRPO)~\citep{grpo} under perception rewards, enabling self-generated reasoning and improved OOD generalization. However, RL-based methods are prone to overthinking due to uncontrolled CoT traces~\citep{pixelthink}.
To curb this overthinking problem, PixelThink~\citep{pixelthink} uses an extra large-scale expert MLLM to estimate query difficulty and cap reasoning length. DR\textsuperscript{2}Seg~\citep{dr2seg} decomposes the task into two passes and rewards a short, self-contained description. 
These methods shorten the reasoning chain, yet leave its fundamental form unchanged: reasoning is still explicitly verbalized in natural language, allowing discrete text tokens and distractor texts to interfere with attention during perception-token generation.

A natural idea is therefore to remove verbalized intermediate reasoning entirely. However, prior studies~\citep{segzero,visionreasoner} have shown that simply removing the reasoning process can degrade model generalization. We therefore explore the potential of latent reasoning~\citep{coconut} in reasoning segmentation. Beyond general latent reasoning in LLMs, latent reasoning segmentation requires latent tokens to jointly encode reasoning semantics and visual-spatial information, while serving as a bridge to generate pixel-level masks, making the task substantially more challenging.

To this end, we propose LIRSeg (\underline{L}atent \underline{I}mplicit \underline{R}easoning for \underline{Seg}mentation), a fully latent reasoning framework for reasoning segmentation. LIRSeg is trained in two stages. 
In the first stage, SFT aligns learnable latent tokens with object-relevant visual regions, grounding latent reasoning in pixel-level evidence.
In the second stage, RL refines these latent tokens using segmentation rewards, enabling latent reasoning within a compact token budget.
Unlike explicit CoT, which inherits rich representational diversity from pretrained MLLMs, latent reasoning with limited data lacks such inherent diversity. We address this limitation with three complementary information-theoretic mechanisms: extreme-advantage sampling to amplify informative learning signals, decoupled exploration–stability updates to preserve disentangled token representations, and latent diversity amplification to prevent representational collapse. As shown in Figure~\ref{fig:teaser}, LIRSeg achieves efficient reasoning and accurate perception without relying on thinking supervision.

Our contributions are summarized as:
\begin{itemize}
\item We propose LIRSeg, the first RL-optimized latent reasoning framework for reasoning segmentation, replacing redundant verbal CoT with learnable latent tokens that refocus MLLMs on task-relevant visual evidence.
\item We propose three complementary mechanisms from an information-theoretic perspective: extreme-advantage sampling, decoupled exploration–stability updates, and latent diversity amplification, which jointly enhance the diversity and representational capacity of latent tokens.
\item We validate the effectiveness of our method across diverse benchmarks and multiple model configurations pairing MLLMs with segmentation models, establishing a new state-of-the-art in efficient reasoning segmentation.
\end{itemize}

\section{Related Work}
\subsection{Reasoning Segmentation}

Referring expression segmentation localizes image regions from direct linguistic descriptions~\citep{lavt,gres,vlt}. Reasoning segmentation extends this setting to implicit queries, where the target need be inferred from contextual cues or semantic relations, requiring both reasoning and visual grounding.

Early approaches for reasoning segmentation primarily rely on SFT, pioneered by LISA~\citep{lisa} and extended along three directions: enhanced token designs and decoders~\citep{pixellm,cores,yang2024lisaimprovedbaselinereasoning,qian2026anchorseglanguagegroundedquery}, multi-round interactive segmentation~\citep{wu2023seesaysegmentteaching,wang2024segllmmultiroundreasoningsegmentation,lu2025rsvpreasoningsegmentationvisual}, and unified visual instruction frameworks~\citep{zhang2024groundhoggroundinglargelanguage,wang2024llmsegbridgingimagesegmentation,wei2024instructsegunifyinginstructedvisual}. However, their reasoning ability is largely inherited from annotated supervision, limiting generalization to unseen scenarios~\citep{segzero}.
More recently, RL has emerged as an alternative, pioneered by Seg-Zero~\citep{segzero} and extended by follow-up works~\citep{visionreasoner,samr1,zhu2025lenslearningsegmentunified,you2025segr1segmentationsurprisinglysimple,zhao2026conceptsegr1segmentconceptmetareinforcement}. However, these methods still generate explicit chain-of-thought (CoT) before producing the mask, incurring substantial overhead and attention interference. Although PixelThink~\citep{pixelthink} and DR$^2$Seg~\citep{dr2seg} adopt adaptive strategies, they still rely on explicit textual CoT, leaving latent reasoning for reasoning segmentation largely unexplored. This work demonstrates the feasibility of this paradigm and broadens its methodological scope.

\subsection{Latent Reasoning}

Latent reasoning has been explored in LLMs through additional computation positions~\citep{pausetoken}, internal rationales~\citep{quietstar}, and continuous-state feedback~\citep{coconut,deng2024implicitcot}. Recent reasoning MLLMs further show that continuous representations can effectively preserve perceptual and reasoning information that is difficult to express in discrete language~\citep{pham2025mcout,yang2025mirage,li2025lvr}.

Prior work provides both methodological and theoretical foundations for exploring this paradigm in reasoning segmentation. This work further investigates the unique challenges of latent reasoning in this setting. Unlike general MLLM reasoning, reasoning segmentation requires latent tokens to jointly encode semantic reasoning and visual-spatial information, while also serving as an interface for generating pixel-level masks. This tighter coupling between reasoning and dense prediction makes optimization substantially more challenging. These considerations motivate the design of LIRSeg, which achieves both accurate and efficient segmentation across diverse benchmarks and different combinations of MLLM backbones and segmentation models.

\begin{figure*}[h!]
\centering
\includegraphics[width=0.98\textwidth]{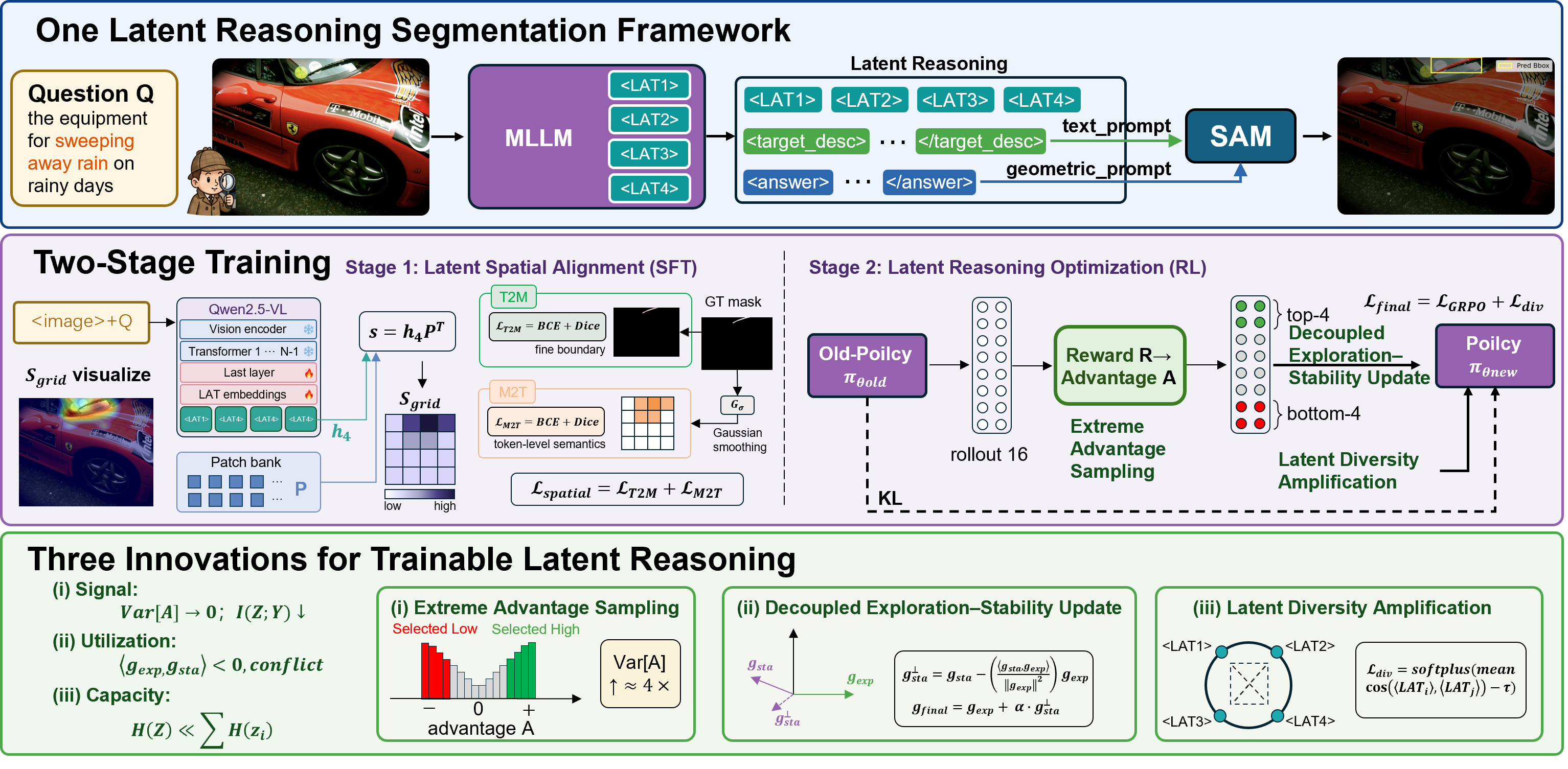}
\caption{\textbf{Overview of LIRSeg.} \emph{Top:} Latent reasoning via latent tokens produces a text description and bounding-box. \emph{Middle:} Two-stage training: spatial alignment followed by RL optimization. \emph{Bottom:} Three innovation components.}
\label{fig:framework}
\end{figure*}

\section{Method}


\subsection{Preliminaries on Reasoning Segmentation}
Reasoning segmentation takes an image $I$ and an implicit textual query $Q$ as input and predicts a binary mask $M$ for the target. Unlike referring segmentation, $Q$ typically does not name the target explicitly, requiring the model to infer the intended object before localization.
Existing methods typically follow an explicit chain-of-thought (CoT) paradigm. Given $(I,Q)$, the MLLM is queried and decodes a language reasoning sequence $\mathcal{T}=(t_1,\ldots,t_L)$ to analyze visual and contextual cues, and then generates structured output $\mathcal{F}_{cot}$ in the format:\textbf{\seqsplit{\texttt{<think>}$\mathcal{T}$\texttt{</think><answer>}$\mathcal{A}$\texttt{</answer>}}} where $\mathcal{T}$ denotes the explicit CoT and $\mathcal{A}$ denotes the perception answer. This prompt is subsequently fed into the segmentation model $\mathcal{S}$ to produce the mask prediction $M$.
\subsection{LIRSeg: A Latent Reasoning Framework for Reasoning Segmentation}
\label{sec:pipeline0}

As shown in Figure~\ref{fig:framework}, we extend the vocabulary of the MLLM with $K$ new special tokens, denoted by
$\langle\text{LAT}_1\rangle,\ldots,\langle\text{LAT}_K\rangle$.
Their token embeddings are learnable and are initialized before training.
The decoding prompt specifies a fixed output template that requires the model to generate these tokens consecutively at the beginning of its response, before producing the target description $\mathcal{D}$ and the perception answer $\mathcal{A}$: 
$\mathcal{F}_{lat}=$
$\langle\text{LAT}_1\rangle \cdots \langle\text{LAT}_K\rangle$\textbf{\seqsplit{\texttt{<target\_desc>}$\mathcal{D}$\texttt{</target\_desc><answer>}$\mathcal{A}$\texttt{</answer>}}}.
Although $\langle\text{LAT}_i\rangle$ are discrete vocabulary tokens, their contextualized final-layer hidden states
$\mathbf{h}_{\mathrm{lat}}^i$
serve as continuous latent reasoning representations.
During autoregressive generation, each latent state attends to the image, the query, and all preceding latent-token positions.
Consequently, the final state $\mathbf{h}_{\mathrm{lat}}^K$ aggregates information from the entire latent-token sequence.
The subsequently generated description $\mathcal{D}$ and answer $\mathcal{A}$ can attend to all preceding latent states through causal self-attention. At segmentation time, we provide SAM3 with the text prompt derived from $\mathcal{D}$ and the bounding-box prompt extracted from $\mathcal{A}$ to obtain the final mask.
\subsection{Two-Stage Training Pipeline}
\label{sec:pipeline}

We train LIRSeg in two stages. Stage~1 spatially grounds the latent tokens, providing a stable initialization for reinforcement learning. Stage~2 then optimizes the policy MLLM and the latent tokens with GRPO under segmentation rewards.

\subsubsection{Stage 1: Warm Start via Spatial Alignment.}
Directly applying RL to randomly initialized latent tokens is unstable, as sparse segmentation rewards provide insufficient supervision for learning meaningful latent tokens from scratch. To this end, we introduce latent spatial alignment, which explicitly grounds latent tokens in target-relevant visual regions without distilling verbose teacher-generated reasoning traces.

Let $\mathbf{h}_{\mathrm{lat}}^{K}\in\mathbb{R}^{d}$ denote the hidden state of the $K$-th latent token at the last transformer layer of the MLLM, and let $P\in\mathbb{R}^{N\times d}$ denote the visual patch features produced by the vision encoder. We compute a spatial similarity map as
\begin{equation}
s = \mathbf{h}_{\mathrm{lat}}^{K} P^{\top}, \qquad S_{\mathrm{grid}} = \mathrm{reshape}(s)\in\mathbb{R}^{H\times W},
\end{equation}
where $S_{\mathrm{grid}}$ indicates the visual regions associated with the latent representation.
We supervise $S_{\mathrm{grid}}$ with the ground-truth mask in two  directions. Token-to-mask (T2M) alignment upsamples $S_{\mathrm{grid}}$ to the pixel resolution and matches it to the mask, promoting boundary localization. Conversely, mask-to-token (M2T) alignment downsamples the Gaussian-smoothed mask into a token-level target and aligns it with $S_{\mathrm{grid}}$, injecting spatial semantics into the latent representation. Both directions are optimized with a combination of binary cross-entropy and Dice losses:
\begin{align}
\mathcal{L}_{\mathrm{T2M}} &= \mathrm{BCE}\!\left(\mathrm{up}(S_{\mathrm{grid}}), M\right) + \mathrm{Dice}(\cdot),\\
\mathcal{L}_{\mathrm{M2T}} &= \mathrm{BCE}\!\left(S_{\mathrm{grid}}, \mathrm{down}(G_\sigma \!*\! M)\right) + \mathrm{Dice}(\cdot).
\end{align}
The total loss of Stage~1 is
\begin{equation}
\mathcal{L}_{\mathrm{spatial}} = \mathcal{L}_{\mathrm{T2M}} + \mathcal{L}_{\mathrm{M2T}}.
\end{equation}
Only the last transformer layer of the MLLM and the learnable latent-token embeddings are updated; all remaining backbone parameters are frozen. This stage provides the latent representation with a spatially meaningful initialization, allowing Stage~2 to refine object-grounded tokens rather than optimize from random noise.

\subsubsection{Stage 2: RL and Reward Design.}
In Stage~2, we optimize the policy using GRPO~\citep{grpo}. Given an input $x=(I,Q)$, we sample a group of $n$ responses $\{y_1,\dots,y_n\}\sim\pi_{\theta_{\mathrm{old}}}$ and assign each response a reward $r_i$. The group-relative advantage is computed as
\begin{equation}
A_i = \frac{r_i-\bar{r}}{\sigma_r+\epsilon},\qquad \bar{r}=\tfrac{1}{n}\sum_j r_j .
\end{equation} 
where $\sigma_r$ is the standard deviation within the group.

The reward function jointly optimizes segmentation quality, localization accuracy, conciseness, and format:
\begin{equation}
r = r_{\text{mask}} + r_{\text{bbox}} 
  + r_{\text{len}} + r_{\text{fmt}}.
\end{equation}
The first two terms provide task-specific supervision. To encourage latent reasoning to encode localization cues without relying on bounding-box coordinates, we evaluate the target description $\mathcal{D}$ by its ability to guide a segmentation model:
\begin{equation}
r_{\text{mask}} = \text{IoU}\bigl(\text{Mask}(I, \mathcal{D}), M_{\text{gt}}\bigr),
\end{equation}
where $\operatorname{Mask}(I,\mathcal{D})$ is the predicted mask and $M_{\text{gt}}$ is the ground-truth mask. Direct localization is supervised using standard bounding-box IoU:
\begin{equation}
r_{\text{bbox}} = \text{IoU}\bigl(B_{\text{pred}}, B_{\text{gt}}\bigr).
\end{equation}
The remaining terms prevent overly verbose or malformed outputs. We reward concise descriptions with
\begin{equation}
r_{\text{len}} = \mathbf{1}_{[1,4]}(|\mathcal{D}|),
\end{equation}
where $|\mathcal{D}|$ denotes the word count. We also introduce a format reward $r_{\text{fmt}}$ to check whether the output conforms to the predefined format $\mathcal{F}_{\mathrm{lat}}$.

\subsection{Optimization for Learnable Latent Tokens}
\label{sec:components}

\subsubsection{Signal: Extreme Advantage Sampling}
\label{sec:extreme}

Replacing verbose CoT with latent tokens reduces rollout diversity because these tokens are insufficiently pretrained to support diverse reasoning trajectories. Consequently, rewards become tightly clustered and advantages concentrate near zero, yielding weak and noise-sensitive policy-gradient signals.
We therefore discard the near-zero middle and estimate the gradient using only the two tails:
\begin{equation}
\widehat{g}_{\mathrm{ext}}
=\frac{1}{K_h+K_l}\sum_{i=1}^{n}
m_i A_i\nabla_\theta\log\pi_\theta(y_i\mid x),
\end{equation}
$m_i=\mathbf{1}\!\left[
A_i\le A_{(K_l)}
\ \vee\
A_i\ge A_{(n-K_h+1)}
\right]$, where $A_{(j)}$ is the $j$-th order statistic. Thus, near-zero advantages yield no update, while extreme advantages give clear directions to reinforce or suppress behaviors.

\subsubsection{Utilization: Decoupled Exploration–Stability Update}
\label{sec:ortho}
The generated output tokens serve two complementary roles in $\mathcal{D}$ and $\mathcal{A}$. Tokens in $\mathcal{D}$ and $\mathcal{A}$ guide the model’s exploration toward the target and are optimized by the IoU reward, inducing an exploration gradient $\mathbf{g}_{\mathrm{exp}}$.
Structural tokens, such as the \texttt{<answer>} tag, are optimized by a format reward and induce a stability gradient $\mathbf{g}_{\mathrm{sta}}$. These gradients exhibit negative cosine similarity, such that naive aggregation can attenuate the reward-aligned exploration signal.

We address this conflict with a decoupled update: we preserve $\mathbf{g}_{\mathrm{exp}}$ and project $\mathbf{g}_{\mathrm{sta}}$ onto the subspace orthogonal to the exploration direction:
\begin{equation}
\mathbf{g}_{\mathrm{sta}}^{\perp} = \mathbf{g}_{\mathrm{sta}} - \frac{\langle \mathbf{g}_{\mathrm{sta}}, \mathbf{g}_{\mathrm{exp}}\rangle}{\lVert \mathbf{g}_{\mathrm{exp}}\rVert^2+\varepsilon}\,\mathbf{g}_{\mathrm{exp}}.
\end{equation}
The final update is
$\mathbf{g} = \mathbf{g}_{\mathrm{exp}} + \mathbf{g}_{\mathrm{sta}}^{\perp}.$
By construction, the cosine similarity $\langle \mathbf{g}_{\mathrm{exp}}, \mathbf{g}_{\mathrm{sta}}^{\perp}\rangle = 0$. Thus, the reward-aligned exploration gradient is well-preserved, while the stability gradient contributes only through orthogonal, non-interfering directions that maintain output structure.

\subsubsection{Capacity: Latent Diversity Amplification}
\label{sec:div}
Although the $K$ registered special tokens have distinct token identities, their contextual hidden states may collapse toward similar directions in the absence of explicit regularization. Such collapse makes the latent representation redundant and limits the information available for predicting $Y=(\mathcal{D},\mathcal{A})$. Let $z_i := \mathbf{h}_{\mathrm{lat}}^i$ denote the final-layer contextual hidden state associated with $\langle\text{LAT}_i\rangle$, and define
$\mathcal{Z}=\{z_1,\ldots,z_K\}$. Their mean pairwise cosine similarity as 
\begin{equation}
\bar{s}=\tfrac{2}{K(K-1)}\sum\cos(z_i,z_j), 
\end{equation}
and introduce the diversity regularizer:
\begin{equation}
\mathcal{L}_{\mathrm{div}} = \operatorname{softplus}\!\big(\bar{s}-\tau_{\mathrm{div}}\big),
\end{equation}
where $\tau_{\mathrm{div}}$ is a similarity threshold. The penalty remains weak when the latent states are sufficiently dispersed and grows approximately linearly as they become aligned. By penalizing redundant latent directions, $\mathcal{L}_{\mathrm{div}}$ helps preserve representational rank and encourages different tokens to capture complementary aspects of the reasoning process.

\begin{table*}[t]
\centering
\small
\caption{\textbf{Performance comparison on ReasonSeg.} We report reasoning tokens to measure efficiency. * marks models trained on the VisionReasoner dataset. Bold and underlined denote best and second-best results, respectively.}
\label{tab:main}
\begin{tabular}{llllcccccc}
\toprule
& & & & \multicolumn{3}{c}{ReasonSeg Val} & \multicolumn{3}{c}{ReasonSeg Test}\\
\cmidrule(lr){5-7}\cmidrule(lr){8-10}
Method & Venue & Language Model & Seg. & Tokens\,\down & gIoU\,\up & cIoU\,\up & Tokens\,\down & gIoU\,\up & cIoU\,\up\\
\midrule
OVSeg       & CVPR'23 & CLIP ViT-L      & -- & --   & 28.5 & 18.6 & --   & 26.1 & 20.8\\
ReLA        & CVPR'23 & BERT            & -- & --   & 22.4 & 19.9 & --   & 21.3 & 22.0\\
\midrule
LISA   & CVPR'24 & LLaVA1.5-7B     & SAM1 & --   & 61.3 & 62.9 & --   & 55.6 & 56.9\\
CoReS & ECCV'24 & LLaVA-7B        & SAM1 & --   & 59.4 & --   & --   & 52.4 & --\\
\midrule
Seg-Zero          & arxiv & Qwen2.5-VL-7B & SAM2 & 90.7 & 61.6 & 52.5 & 90.6 & 58.2 & 52.3\\
SAM-R1              & NeurIPS'25 & Qwen2.5-VL-7B & SAM2 & --   & 64.0 & 55.8 & --   & 60.2 & 54.3\\
PixelThink     & ICLR'26 & Qwen2.5-VL-7B & SAM2 & 46.9 & 63.8 & 62.6 & 47.6 & 60.1 & 55.7\\
VisionReasoner$^{*}$   & ICLR'26 & Qwen2.5-VL-7B & SAM3 & 80.8 & 66.3 & 59.8 & 84.8 & 63.6 & 58.2\\
VisionReasoner         & ICLR'26 & Qwen2.5-VL-7B & SAM2 & 85.3 & 65.4 & 60.3 & 81.4 & 62.3 & 54.6\\
VisionReasoner         & ICLR'26 & Qwen2.5-VL-7B & SAM3 & 64.8 & 65.8 & 61.5 & 64.7 & 65.5 & 59.2\\
DR$^2$Seg$^{*}$       & ICML'26 & Qwen2.5-VL-7B & SAM2 & 46.2 & 67.5 & 60.0 & 55.4 & 64.8 & 62.8\\
DR$^2$Seg             & ICML'26 & Qwen2.5-VL-7B & SAM2 & \underline{26.9} & 68.5 & 65.8 & \underline{27.2} & 66.1 & \underline{63.6}\\
DR$^2$Seg             & ICML'26 & Qwen2.5-VL-7B & SAM3 & 31.2 & \underline{69.4} & \underline{66.4} & 30.7 & \underline{66.5} & 61.7\\
\midrule
\textbf{LIRSeg (Ours)$^{*}$} & -- & Qwen2.5-VL-7B & SAM3 & \best{4} & 69.0 & 60.6 & \best{4} & 65.2 & 63.1\\
\textbf{LIRSeg (Ours)      } & -- & Qwen2.5-VL-7B & SAM3 & \best{4} & \best{71.7} & \best{67.7} & \best{4} & \best{70.4} & \best{63.8}\\
\bottomrule
\end{tabular}
\end{table*}

\subsection{Theoretical Motivation}
\label{sec:theory}

\paragraph{Verbalization Bottleneck.}
LIRSeg retains reasoning in continuous latent states $\mathcal{Z}$ instead of verbalizing it into a discrete token sequence $\mathcal{T}$ before mask construction. Consider matched upstream computation in which $\mathcal{T}\sim q(\mathcal{T}\mid\mathcal{Z})$ is the reasoning interface passed downstream. Then $M-\mathcal{Z}-\mathcal{T}$ forms a Markov chain: given the latent state, the text tokens are conditionally independent of the target mask. The data-processing inequality gives $I(M;\mathcal{T}) \leq I(M;\mathcal{Z})$, or equivalently
\begin{equation}
I(M;\mathcal{Z})-I(M;\mathcal{T})
=I(M;\mathcal{Z}\mid\mathcal{T}) \geq 0.
\label{eq:latent-bottleneck}
\end{equation}
Verbalization is therefore lossless only when $\mathcal{T}$ is a sufficient statistic of $\mathcal{Z}$ for predicting $M$. Otherwise, task-relevant information in the latent state is discarded by the textual interface. This result does not assume that continuous representations are intrinsically more informative; it only motivates preserving the latent reasoning state rather than forcing an intermediate verbalization.

\paragraph{Latent sufficiency.}
Because LIRSeg decodes the mask directly from $\mathcal{Z}$, the residual uncertainty is $H(M\mid\mathcal{Z})$. Had one instead used the textual interface, the chain rule for entropy together with the Markov property $H(M\mid\mathcal{Z},\mathcal{T})=H(M\mid\mathcal{Z})$ yields
\begin{equation}
H(M\mid\mathcal{T})
=H(M\mid\mathcal{Z})+I(M;\mathcal{Z}\mid\mathcal{T}),
\label{eq:sufficiency}
\end{equation}
which implies $H(M\mid\mathcal{T}) \ge H(M\mid\mathcal{Z})$.
The inequality is strict whenever $\mathcal{T}$ is not sufficient for $\mathcal{Z}$. Thus the latent state is never less informative than its verbalization; preserving $\mathcal{Z}$ avoids the additional prediction uncertainty introduced by discrete compression.

\paragraph{Compact attention.}
Explicit CoT expands to a long sequence $\mathcal{T}$ with $|\mathcal{T}|\gg|\mathcal{Z}|$, whose tokens compete with visual features for attention capacity. Combining Eq.~\eqref{eq:latent-bottleneck} with the token budget gives the per-token task information
\begin{equation}
\frac{I(M;\mathcal{Z})}{|\mathcal{Z}|}
\;\gg\;
\frac{I(M;\mathcal{T})}{|\mathcal{T}|}.
\label{eq:density}
\end{equation}
Because $\mathcal{Z}$ is compact and interacts directly with visual features without intermediate language, it avoids the attention dilution and increased effective distance between perception tokens caused by redundant text. Each latent token therefore participates in undiluted visual attention, whereas language tokens insert non-visual keys and values that lengthen the reasoning path and introduce irrelevant image distractors.

\section{Experiments}

\subsection{Experimental Settings}

\paragraph{Datasets.} 
For training, we use either the ReasonSeg training set (only 239 samples) or the VisionReasoner-7K dataset separately, to examine generalization under different training data. We evaluate on three reasoning segmentation benchmarks and the RefCOCO referring suite. ReasonSeg tests multi-instance reasoning, MUSE tests multi-target semantic discrimination, MMR requires joint object- and part-level masks, and RefCOCO verifies generalization.

\paragraph{Evaluation Metrics.} 

Following prior work~\citep{lisa,dr2seg}, we report gIoU and cIoU for accuracy, and the average number of reasoning tokens for efficiency.

\paragraph{Experimental Details.} 
All experiments are conducted on 4 H100 GPUs with a per-device batch size of 8. By default we use Qwen2.5-VL-7B~\citep{qwen25vl} and SAM3~\citep{sam3}. We set $K=4$ latent tokens. Stage~1 performs SFT for 500 steps. Stage~2 applies RL for 5 epochs with $n=16$ rollouts, $K_h=K_l=4$ extreme sampling.

\begin{figure*}[ht!]
    \centering
    \includegraphics[width=0.98\linewidth]{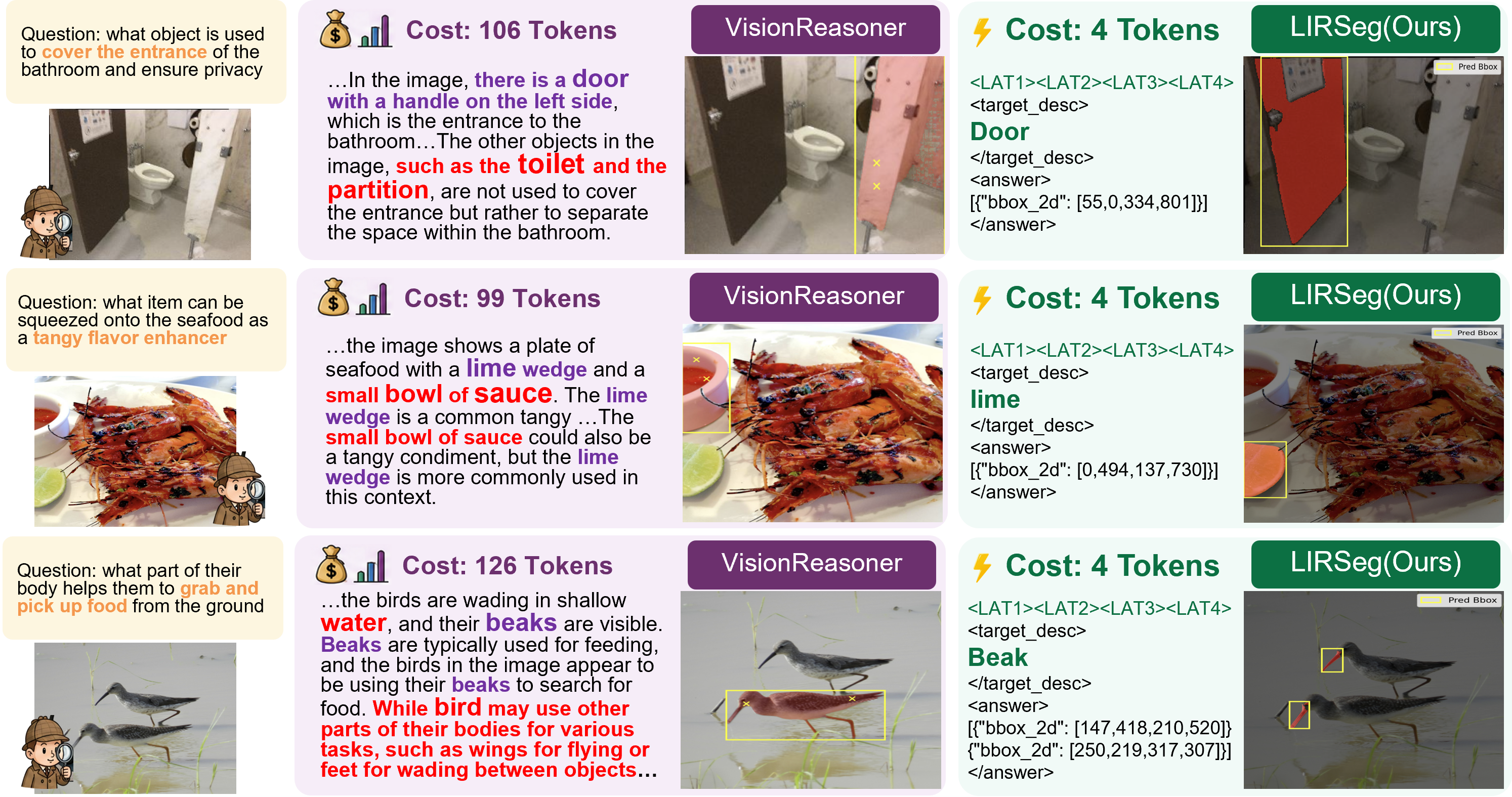}
    \caption{\textbf{Qualitative comparisons between VisionReasoner and our LIRSeg.} The representative samples are selected from simple single-object to complex multi-object scenarios, which reveals severe attention interference.}
    \label{fig:qual}
\end{figure*}

\subsection{Main Results}

\paragraph{Comparison Methods.} 
We compare LIRSeg with non-MLLM segmentation methods
~\citep{ovseg,liu2023gresgeneralizedreferringexpression},
SFT-based MLLM methods~\citep{lisa,cores,xia2024gsvageneralizedsegmentationmultimodal,rasheed2024glammpixelgroundinglarge,mmr},
explicit-CoT RL methods~\citep{segzero,popen,samr1,pixelthink,visionreasoner,dr2seg}.
Finally, our LIRSeg, a latent reasoning method, replaces explicit CoT with a compact set of learnable latent tokens, enabling efficient encoding of task-relevant visual and semantic information.

\subsubsection{Reasoning Segmentation Results}

\paragraph{Multi-Target Single-Semantic Scenes (ReasonSeg).} 

Table~\ref{tab:main} reports results on ReasonSeg. Under the same Qwen2.5-VL-7B backbone and SAM3 segmenter, LIRSeg achieves 71.7/67.7 gIoU/cIoU on val and 70.4/63.8 on test, surpassing VisionReasoner by 5.9/6.2 and 4.9/4.6, respectively. LIRSeg reduces reasoning tokens by ~16× (64.8 → 4). Per-sample inference time drops from 0.783 s to 0.512 s, achieving 53\% speedup over VisionReasoner. It also outperforms the recent state-of-the-art method DR$^2$Seg by 2.3 and 3.9 gIoU on val/test. These results demonstrate that latent reasoning effectively alleviates attention interference, improving both accuracy and efficiency. Latent reasoning also generalizes well in the zero-shot setting: LIRSeg$^*$, trained only on the VisionReasoner-7K dataset (short referring expressions without complex reasoning) and directly evaluated on ReasonSeg, improves over VisionReasoner$^*$ by 2.7 gIoU on val while reducing tokens by 20$\times$ (80.8 $\rightarrow$ 4). This suggests that latent reasoning is not task-specific overfitting; rather, its compact bottleneck directly encodes task-relevant visual evidence, enabling transfer from simple to complex queries.

\begin{table}[h!]
\centering
\small
\caption{\textbf{Results on MUSE and MMR benchmarks.} Performance is reported in terms of gIoU. A Dash indicates that the corresponding method did not report results.}
\label{tab:musemmr}
\begin{tabular}{l|c|c|c|c}
\toprule
& \multicolumn{2}{c|}{\textbf{MUSE}} & \multicolumn{2}{c}{\textbf{MMR}} \\
Method & val & test & val & test \\
\midrule
LISA & 42.0 & 38.9 & 19.4 & 19.5 \\
PixelLM & 42.6 & 39.2 & -- & -- \\
POPEN & 45.4 & 42.4 & -- & -- \\
\midrule
GSVA & -- & -- & 19.8 & 21.2 \\
GLaMM & -- & -- & 26.9 & 30.3 \\
M$^2$SA & -- & -- & \underline{27.8} & \underline{30.9} \\
\midrule
VisionReasoner & \underline{50.5} & \underline{47.7} & 26.7 & 28.4 \\
LIRSeg         & \best{55.8} & \best{54.8} & \best{29.6} & \best{33.1} \\
\bottomrule
\end{tabular}
\end{table}

\paragraph{Multi-Target Multi-Semantic Scenes (MUSE).}

Table~\ref{tab:musemmr} (left) reports MUSE results, where each query describes multiple targets with diverse semantics. LIRSeg outperforms VisionReasoner by 5.3 and 7.1 gIoU on val and test splits, respectively. These gains show that latent tokens effectively encode multi-target semantics while avoiding the bottleneck and attention interference of explicit text reasoning.

\paragraph{Multi-Target Multi-Granularity Scenes (MMR).}
Table~\ref{tab:musemmr} reports results on MMR, which requires segmentation at both the object and part levels. LIRSeg outperforms VisionReasoner by 2.9 gIoU on the val split and 4.7 gIoU on the test split. The consistent gains demonstrate that latent reasoning remains effective across varying levels of granularity.

\begin{table}[h!]
\centering
\small
\caption{\textbf{Referring segmentation, RefCOCO series.}}
\label{tab:res}
\begin{tabular}{lccc}
\toprule
Method & RefCOCO & RefCOCO+ & RefCOCOg\\
       & testA   & testA    & test\\
\midrule
LISA         & 76.5 & 67.4 & 68.5\\
Seg-Zero  & \best{80.3} & \best{76.2} & \underline{72.6}\\
VisionReasoner & \underline{78.8} & 75.1 & 71.5\\
DR$^2$Seg      & 78.7 & \underline{75.4} & 72.2\\
\midrule
LIRSeg         & 78.7 & 75.0 & \best{72.7}\\
\bottomrule
\end{tabular}
\end{table}

\subsubsection{Referring Segmentation Results}

Table~\ref{tab:res} evaluates generalization on the RefCOCO series, on which LIRSeg achieves 78.7/75.0/72.7~gIoU, attaining the best result on RefCOCOg while remaining competitive on the others. This demonstrates that latent reasoning also preserves strong performance in simple referring scenarios.

\subsubsection{Qualitative Results}

Figure~\ref{fig:qual} provides a qualitative comparison between VisionReasoner and LIRSeg. LIRSeg produces concise, target-specific outputs and consistently localizes the intended object, even in cluttered scenes. In contrast, the explicit-reasoning baseline is more prone to drifting toward distractors mentioned in its CoT traces.

\subsection{Diagnostic Experiments}

\begin{figure}[t]
    \centering
    \includegraphics[width=0.98\linewidth]{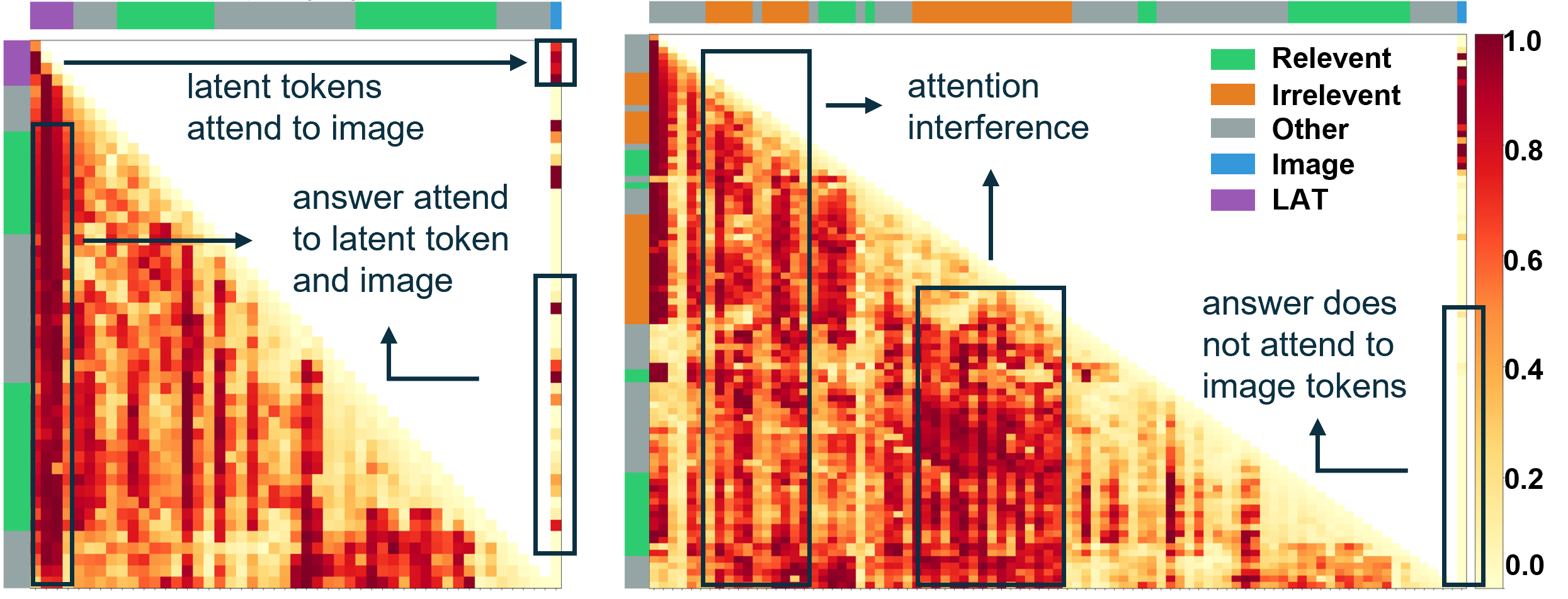}
    \caption{\textbf{Attention confusion phenomenon.} Left: latent reasoning. Right: explicit reasoning.}
    \label{fig:attn}
\end{figure}

\subsubsection{Ablation on Core Components}
\begin{table}[t]
\centering
\small
\caption{\textbf{Component ablation of LIRSeg.} EAS: extreme advantage sampling; DES: decoupled exploration–stability update; LDA: latent diversity amplification.}
\label{tab:ablation}
\begin{tabular}{lccc}
\toprule
Configuration & Tokens\,\down & gIoU\,\up & cIoU\,\up\\
\midrule
\textbf{LIRSeg (full)}         & 4 & \best{71.7} & \best{67.7}\\
\quad w/o Stage~1 alignment    & 4 & 68.6 & 60.9\\
\quad w/o EAS                  & 4 & 69.2 & 61.0\\
\quad w/o DES                  & 4 & 69.9 & 59.4\\
\quad w/o LDA                  & 4 & 68.2 & 60.1\\
\bottomrule
\end{tabular}
\end{table}


Table~\ref{tab:ablation} shows leave-one-out ablations on ReasonSeg validation set, where removing any component consistently drops gIoU by 2–3 points. In particular, ablating LDA causes the largest decline (68.2 gIoU), and the consistent degradation across all variants clearly underscores their complementary synergy in boosting both segmentation accuracy and robustness, further confirming that each component targets a distinct informational bottleneck.

\subsubsection{Empirical Analysis of Attention Interference}
Figure~\ref{fig:attn} compares attention patterns under explicit and latent reasoning. In the explicit-reasoning model (right), the output token attends broadly over the generated reasoning chain, assigning attention to irrelevant text tokens (orange) while weakening its focus on image tokens (blue) and relevant text (green). This pattern illustrates attention interference, where redundant verbal CoT disrupts visual grounding. In contrast, the latent-reasoning model (left) concentrates latent-token attention (purple) on task-relevant image regions. The answer token then attends directly to the image and latent tokens, avoiding interference from irrelevant textual reasoning.

\begin{table}[h]
\centering
\small
\caption{\textbf{Stratified comparison by reasoning difficulty.} Samples are split at the median think-token count of the explicit baseline. $\Delta$gIoU denotes the absolute improvement of LIRSeg over the explicit baseline.}
\label{tab:stratified}
\begin{tabular}{lcccccc}
\toprule
Group & Threshold & Samples & Explicit & Latent & $\Delta$gIoU\\
\midrule
Overall & -- & 200 & 65.8 & 71.7 & +5.9\\
Easy & $\le$68 & 102 & 67.5 & 70.6 & +3.1\\
Hard & $>$68 & 98 & 64.0 & 72.9 & \best{+8.9}\\
\bottomrule
\end{tabular}
\end{table}

\subsubsection{Impact on Easy and Hard Reasoning Cases}

We use the median reasoning tokens from VisionReasoner to divide samples into easy and hard groups. As shown in Table~\ref{tab:stratified}, LIRSeg improves gIoU by 8.9 on hard samples, compared to 3.1 on easy samples. This larger improvement suggests that latent reasoning goes beyond compact visual anchoring by capturing task-relevant compositional information whose importance increases with reasoning complexity. The results also support the attention-interference analysis in Figure~\ref{fig:attn}, indicating that removing explicit reasoning is particularly beneficial for queries requiring longer inference chains.

\begin{table}[h!]
\centering
\small
\caption{\textbf{Ablation on Number of latent tokens.}}
\label{tab:latent}
\begin{tabular}{lcc}
\toprule
$K$ & gIoU\,\up & cIoU\,\up\\
\midrule
$0$          & 67.6 & 55.4\\
$1$          & 69.1 & 59.8\\
$2$          & 69.5 & 64.9\\
\textbf{$4$}   & \best{71.7} & \best{67.7}\\
$8$          & 68.1 & 58.9\\
\bottomrule
\end{tabular}
\end{table}

\subsubsection{Effect of Latent Token Number}

Table~\ref{tab:latent} ablates the latent token count $K$. A single token ($K=1$) underperforms 69.1 gIoU due to insufficient capacity, while $K=8$ introduces redundancy and drops to 68.1 gIoU. The best result is at $K=4$.  This confirms a capacity--compression trade-off: too few tokens fail to encode sufficient reasoning, while too many weaken the bottleneck and encourage redundancy.

\subsubsection{Generality across Backbones and Segmenters}
\begin{table}[h!]
\centering
\small
\caption{\textbf{Generality across backbones and segmenters.}}
\label{tab:general}
\begin{tabular}{llcccc}
\toprule
& & \multicolumn{2}{c}{Val} & \multicolumn{2}{c}{Test}\\
\cmidrule(lr){3-4}\cmidrule(lr){5-6}
MLLM & Seg. & gIoU & cIoU & gIoU & cIoU\\
\midrule
Qwen2.5-VL & SAM2 & 68.8 & 62.6 & 67.8 & 64.5\\
Qwen2.5-VL & SAM3 & 71.7 & 67.7 & 70.4 & 63.8\\
Qwen3-VL   & SAM2 & \best{73.2} & \best{67.9} & \best{71.0} & \best{68.7}\\
Qwen3-VL   & SAM3 & 72.1 & 62.8 & 70.8 & 64.4\\
\bottomrule
\end{tabular}
\end{table}

Table~\ref{tab:general} evaluates LIRSeg across different MLLM backbones and segmenters. All configurations outperform their explicit-reasoning counterparts on ReasonSeg, with Qwen3-VL~\citep{qwen3vl} + SAM2 achieving the best result of 73.2/71.0 gIoU on the val/test splits. These results demonstrate the strong architectural generalization of LIRSeg.

\section{Conclusion}
LIRSeg introduces latent reasoning for reasoning segmentation by replacing explicit CoT with a compact set of learnable latent tokens. These tokens are first grounded in task-relevant visual evidence through spatial alignment and are then optimized with segmentation rewards. To make this compact latent bottleneck effective, we develop three complementary mechanisms: extreme-advantage sampling to strengthen reward signals, decoupled exploration--stability updates to preserve useful gradients, and latent diversity amplification to prevent representational collapse. Across multiple MLLM backbones and segmenters, LIRSeg consistently improves segmentation accuracy while reducing reasoning tokens by approximately $16\times$, achieving state-of-the-art results on ReasonSeg, MUSE, and MMR and remaining competitive on RefCOCO. These results show that explicit verbal reasoning is not necessary for accurate reasoning segmentation and provide new insights into how compact latent computation can improve both visual grounding and reasoning efficiency.

\newpage

\bibliography{aaai2027}

\end{document}